\documentclass{article}

\usepackage{microtype}
\usepackage{graphicx}

\usepackage{subcaption}

\usepackage{graphicx}
\usepackage{subcaption}

\usepackage{booktabs} % for professional tables

\usepackage{hyperref}
\usepackage[numbers,sort]{natbib}

\usepackage[accepted]{icml2025}
\setcitestyle{numbers,square,sort&compress}

\usepackage{amsmath}
\usepackage{amssymb}
\usepackage{mathtools}
\usepackage{amsthm}

\usepackage[capitalize,noabbrev]{cleveref}

\theoremstyle{plain}

\theoremstyle{definition}

\theoremstyle{remark}

\usepackage[textsize=tiny]{todonotes}

\begin{document}

\twocolumn[
\icmltitle{From World Models to World Action Models: A Concise Tutorial for Robotics}



\icmlsetsymbol{equal}{*}

\begin{icmlauthorlist}
\icmlauthor{Xiaoxiong Zhang}{sust},
\icmlauthor{Xiong Zeng}{sust}, and 
\icmlauthor{Wei Zhang}{sust,limx}

\end{icmlauthorlist}

\icmlaffiliation{sust}{School of Automation and Intelligent Manufacturing, Southern University of Science and Technology, Shenzhen, China}
\icmlaffiliation{limx}{LimX Dynamics, Shenzhen, China}

\icmlcorrespondingauthor{Xiaoxiong Zhang}{12433017@mail.sustech.edu.cn}


\vskip 0.3in
]



\printAffiliationsAndNotice{} 
% otherwise use the standard text.

\begin{abstract}
 {Rather than providing an exhaustive survey, this paper presents a concise tutorial on world models and world action models for robotics. After reading the tutorial, readers should have a clear understanding of what constitutes a "world", how world models and world action models are defined, and what roles they play within robotic AI systems. The tutorial also develops a unified perspective for comparing representative approaches, such as World Labs' spatial intelligence models, Yann LeCun's JEPA framework, and NVIDIA's Cosmos platform, and clarifies how these models differ in their representations, predictive capabilities, and interaction mechanisms.}

Github page: \href{https://github.com/clearlab-sustech/WorldModelSurvey}{World (Action) Models}.
\end{abstract}

\section{Foundations: World, Task, Policy, and Models}
\label{sec:introduction}

Predicting how the world evolves under actions is a central capability for physical artificial intelligence (AI) and generative simulation. In these scenarios, the predictive models can support controller design, planning, decision making, and policy learning. Recently, the terms  {world model} and  {world action model} have been used to describe a broad range of methods, including latent dynamics models, video prediction models, physics-informed simulators, and action-conditioned generative models. However, these methods differ substantially in what they model, what they predict, and how their predictions are used. This motivates a unified definition and taxonomy of world (action) models.

\begin{figure}[!htbp]
    \centering
    \includegraphics[width=0.7\linewidth]{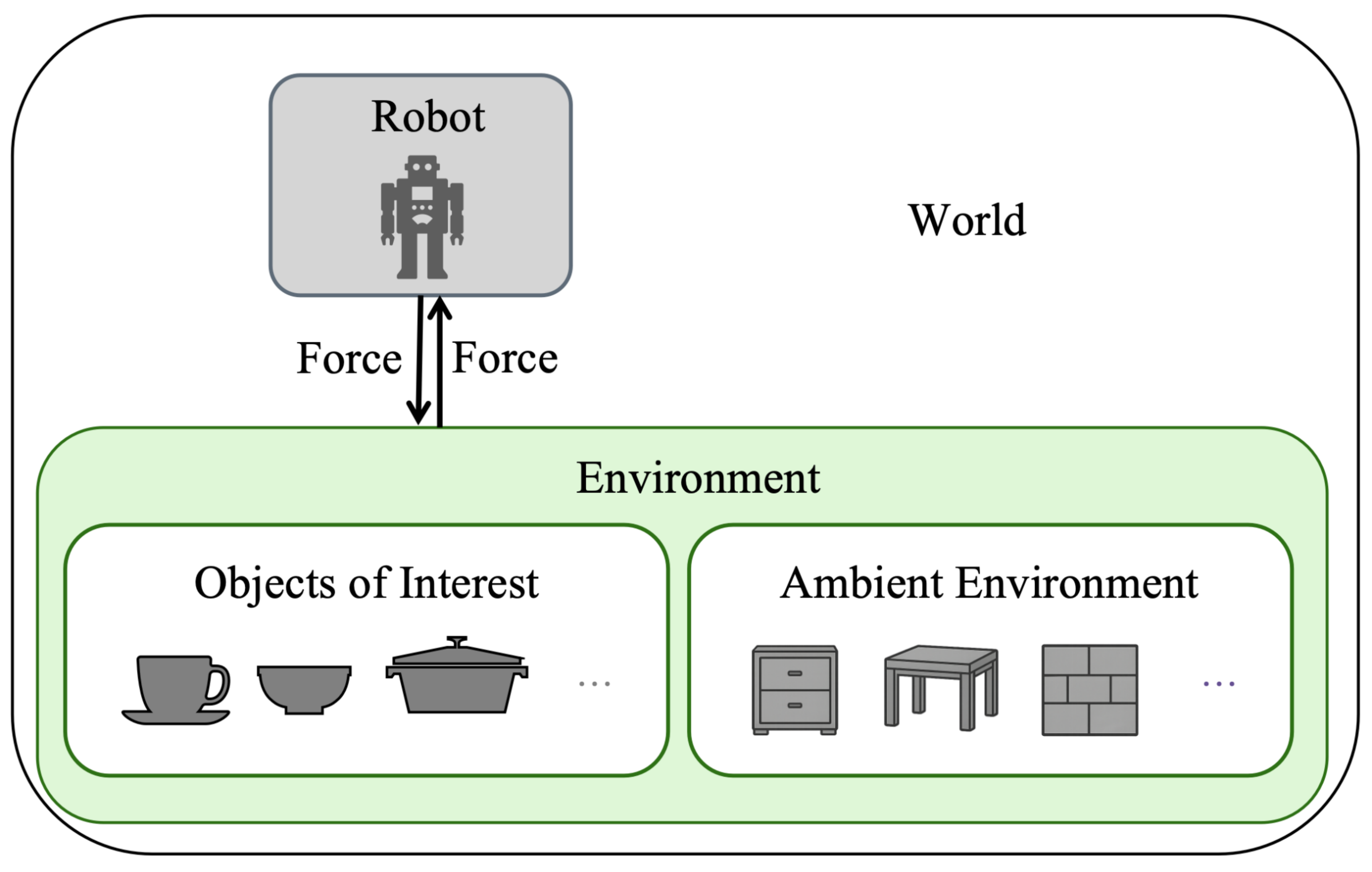}
     \caption{Illustration of the components of a world.}

    \label{fig:world_definition}
\end{figure}

\subsection{World}

We define a \emph{world} as the set of task-relevant entities, including both the \emph{robot} and its \emph{environment}.
The environment contains the \emph{objects of interest} and the  \emph{ambient environment}, as shown in Fig. \ref{fig:world_definition}.

\begin{figure}[!htbp]
    \centering
    \includegraphics[width=0.7\linewidth]{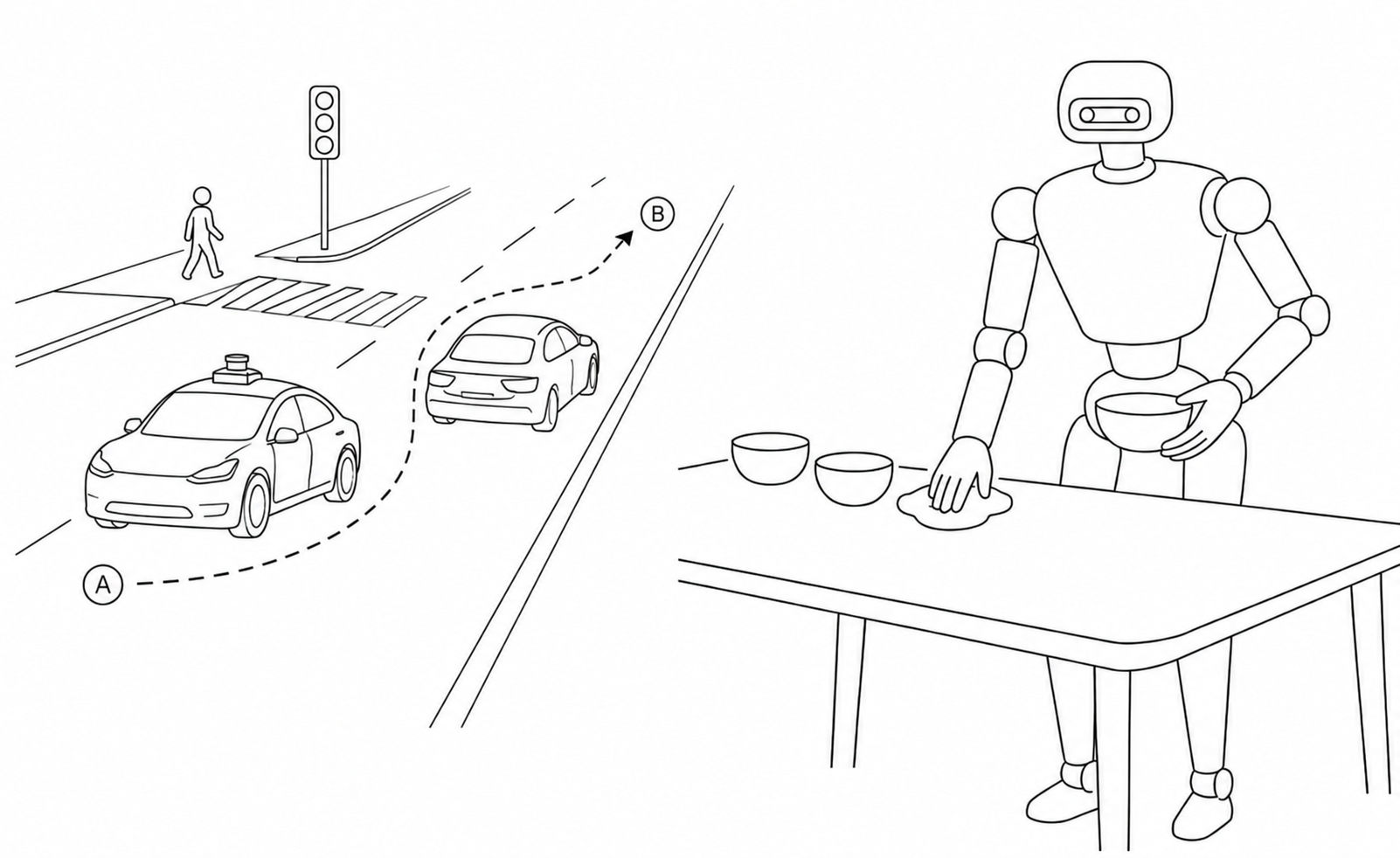}
    \caption{{Physical AI tasks: Autonomous driving and robotic table-cleaning.}}
    \label{fig:two_tasks}
\end{figure}

\subsection{Physical AI Task and Policy}
We define the \emph{world state} as the joint task-relevant state of the robot, objects, and the surrounding environment. A \emph{physical AI task} is to design a policy that drives the world from an initial state toward a goal set while satisfying task-specific constraints. Different tasks induce different choices of relevant entities, state variables, objectives, and constraints. Fig. \ref{fig:two_tasks} provides two examples of task-specific worlds:

\vspace{-3pt}
\noindent \emph{Autonomous driving task.} In the autonomous driving task, the world consists of the vehicle, all relevant traffic participants, and the surrounding environment. The task is to design a driving policy that moves the vehicle from an initial location A to a target location B, while avoiding collision throughout the journey. 

\vspace{-3pt}
 \noindent \emph{Robotic table-cleaning task.} In the robotic table-cleaning task, the world includes the robot and all relevant objects such as dishes, a rag, a table, and a household environment, and the objective is to rearrange objects into acceptable states (e.g., dishes placed in a sink).

\begin{figure}[!htbp]
    \centering
    \includegraphics[width=0.7\linewidth]{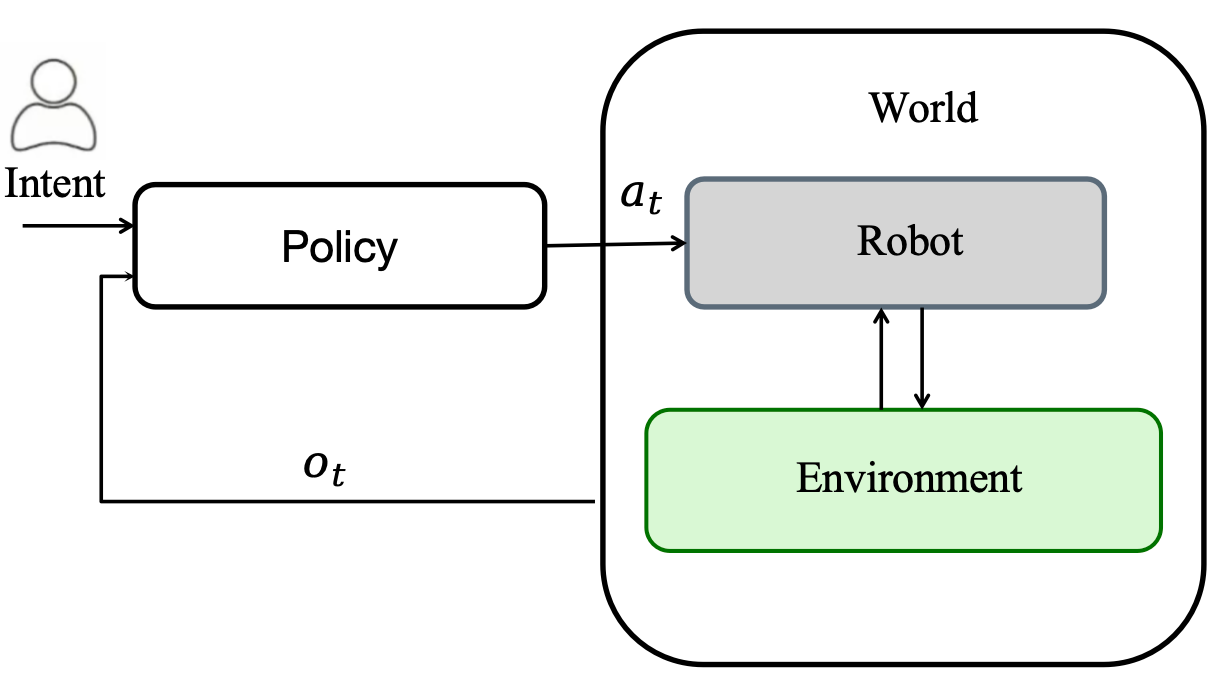}
    \caption{A policy framework for accomplishing physical AI tasks.}
    \label{fig:world_action_model_framework}
\end{figure}

  {Fig. \ref{fig:world_action_model_framework} illustrates
 a policy framework that maps human intent (e.g., driving from A to B in the autonomous driving task) and the current observation \(o_t\) to a robot action \(a_t\). This mapping may be realized by a modular decision–planning–control pipeline, a fast–slow architecture, or an end-to-end policy, depending on the task and the available data. Low-level control modules may use methods such as PID or MPC, whereas learned action-generation components may adopt architectures such as VLA models or world action models. These choices concern different architectural and methodological dimensions rather than different functional definitions of a policy.} 
% that, in an physical AI task, the \emph{policy} receives a language instruction $l$ and the current observation $o_t$
% % from the world, and then outputs an action $a_t$ to the robot. 

% {A policy for accomplishing a physical AI task maps human intent \(I\) (e.g., driving from \(A\) to \(B\)) and the current observation \(o_t\) to a robot action \(a_t\). It may be implemented as a modular decision--planning--control pipeline, a fast--slow architecture, or an end-to-end policy, using approaches such as model predictive control (MPC), vision-language-action (VLA) models, or world action models (WAMs). These model-based, data-driven, and end-to-end approaches are alternative implementations of the same policy functionality.}

%  {To finish an physical AI task,  we need to design a policy that maps from human intent (e.g. A-B in the autonomous driving case), observation to a control action for  the robot.  There are different ways to design such a policy,  sometimes dividing into different modules (decision, plannning, control), or fast-slow system, end-to-end policies if we easily obtain enough end-to-end data.  Depending on the task nature, we could use MPC,  VLA, WAM, ect. Key to note is that whether end-to-end, model based, or data driven, do not change the functionality of the policy, they are merely speficic ways to accomplish a task at hand. }

% The policy might be a proportional-integral-derivative (PID) controller, a model predictive controller (MPC), a vision-language-action (VLA) model, or a world action model (WAM).

 \subsection{World Models and World Action Models}
 \label{subsec_WM_WAM}

%  \begin{figure}[t]
%     \centering
%     \includegraphics[width=0.7\linewidth]{figs/world_model.pdf}
%     \caption{A world model predicts future observations or states from the current observation and the current action. 
%     }
%     \label{fig:overview}
% \end{figure}

% \begin{figure}[t]
%     \centering
%     \includegraphics[width=0.7\linewidth]{figs/world_action_model.pdf}
%     \caption{A world action model predicts future actions (and future observations) from the current observation and the current action. 
%     % \textbf{Left:} an observation-space world model directly predicts the 
%     }
%     \label{fig:wordl_am}
% \end{figure}

\begin{figure}[!htbp]
    \centering
    \begin{subfigure}[t]{0.5\linewidth}
        \centering
        \includegraphics[width=\linewidth]{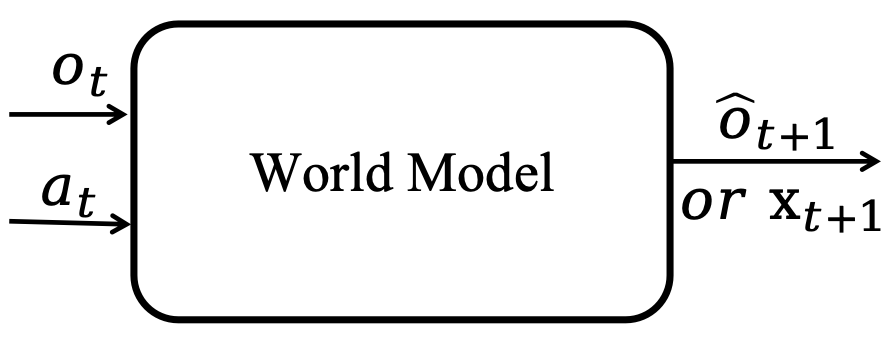}
        % \caption{World model.}
        \label{fig:world_model}
    \end{subfigure}
    \hfill
    \begin{subfigure}[t]{0.5\linewidth}
        \centering
        \includegraphics[width=\linewidth]{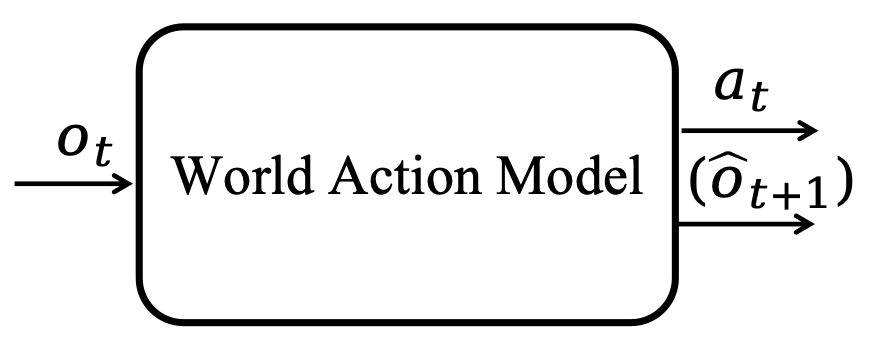}
        % \caption{World action model.}
        \label{fig:world_action_model}
    \end{subfigure}
    \caption{World (action) models.}
    \label{fig:world_model_comparison}
\end{figure}

 For a specified world, a world model predicts how the observation prediction $\hat{o}_{t+1}$ or state prediction $x_{t+1}$ evolve under candidate actions $a_t$, typically conditioned on the current observation $o_{t}$, as illustrated in Fig.~\ref{fig:world_model_comparison}.  

\begin{figure}[!htbp]
    \centering
    \includegraphics[width=0.98\linewidth]{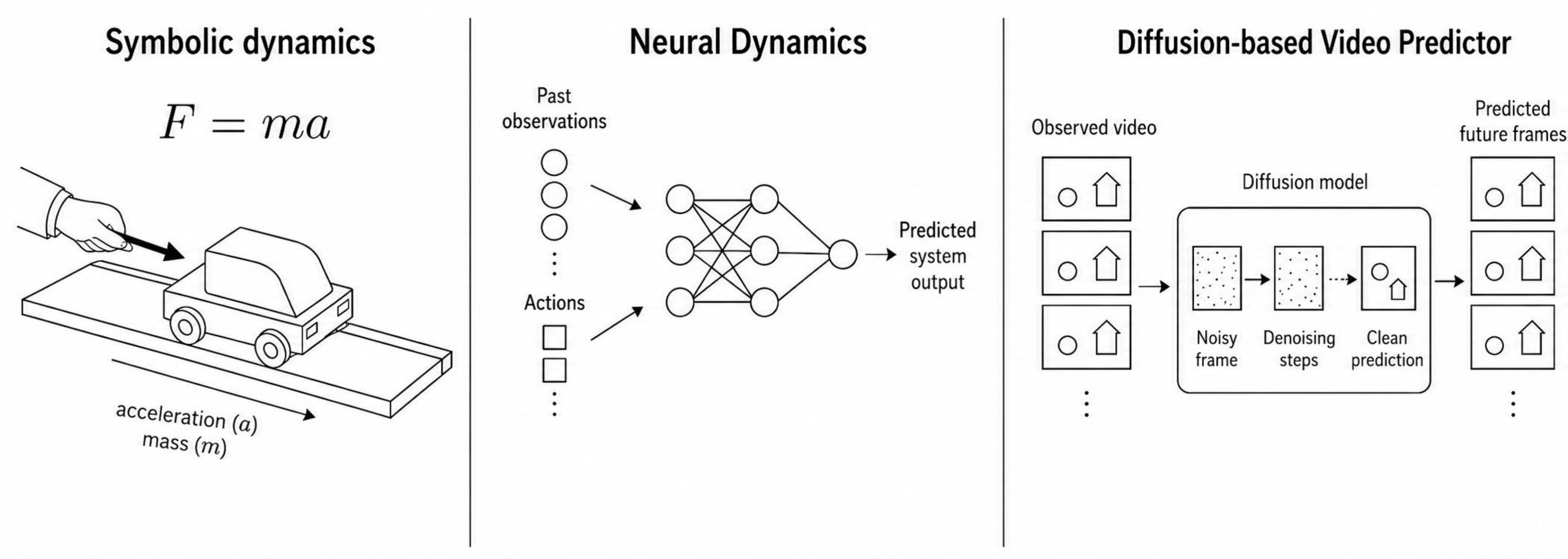}
    \caption{Examples of world models.}
    \label{fig:ex_wm}
\end{figure}

The world model (WM) may take different forms, including symbolic dynamics equations, neural network dynamics models, or diffusion-based video predictors, as shown in Fig. \ref{fig:ex_wm}. 
 {The observation $o_t$ and the observation preidction $\hat{o}_{t+1}$ may comprise the robot's proprioceptive observations as well as environmental observations of the surrounding environment, such as RGB or RGB-D images or point clouds.}
 The action $a_t$ might
range from a concrete robot action to an abstract language-level instruction or a camera pose to observe the environment.
The state prediction
 \(x_{t}\) may represent object poses, object velocities, a latent state, or other variables relevant to the task. We discuss world models in more detail in Sec. \ref{sec:design}. 

% A \emph{world action model} is a policy in the physical AI task. It extends world models by explicitly associating predicted future observations or states with actions, as shown in Fig.~\ref{fig:world_model_comparison} (b).  

% For a given world, a world model predicts a distribution over future observations or states under candidate actions, conditioned on the information available at the current time.

A \emph{world action model} (WAM) is a policy whose action generation is coupled, during inference, with observation prediction, as shown in Fig.~\ref{fig:world_model_comparison}. This coupling may be explicit, latent, joint, or auxiliary. A world action model therefore need not contain a separate world model at inference time.

\noindent \textbf{The differences between WM and WAM} are summarized as follows:

\vspace{-3pt}
\noindent \emph{Primary objective.}
A world model aims to model the evolution of the environment by predicting future observations or states, typically conditioned on the current observation and the current action.
In contrast, a world action model serves as a policy for accomplishing a physical AI task: it predicts actions that are expected to drive the world toward a desired future state or observation.

\vspace{-3pt}
\noindent \emph{System role.}
A world model primarily addresses environment modeling and prediction, whereas a world action model addresses decision-making and control.
Although a world action model may internally use a world model for planning or action evaluation, this is not a necessary architectural requirement.

\vspace{-3pt}
\noindent \emph{Architectural implication.}
A world model does not, by itself, constitute a policy or controller.
To complete a physical AI task, the predictive capability of a world model must therefore be coupled with an additional planning, decision-making, or control mechanism.
 
Our taxonomy of world action models focuses primarily on recent video-based robotic policies, where future visual prediction is coupled with action generation. Classical model-based control and model-based reinforcement learning are discussed only when they clarify this connection.

Further details about world action models are provided in Sec.~\ref{sec_WAM}.

\vspace{-2pt}
\noindent \textbf{The differences between WAM and VLA:} A VLA typically learns a direct mapping from multimodal observations and language instructions to robot actions, emphasizing end-to-end policy learning. In contrast, a WAM explicitly reasons about how actions transform the world toward a desired future state, often leveraging predictive world representations to evaluate, plan, or generate actions. Taking Fig.~\ref{fig:world_model_comparison} as a comparison example, the VLA takes only the action $a_t$ as output and its output does not include the observation prediction $\hat{o}_{t+1}$.

\begin{figure*}[!htbp]
    \centering
    \includegraphics[width=0.8\linewidth]{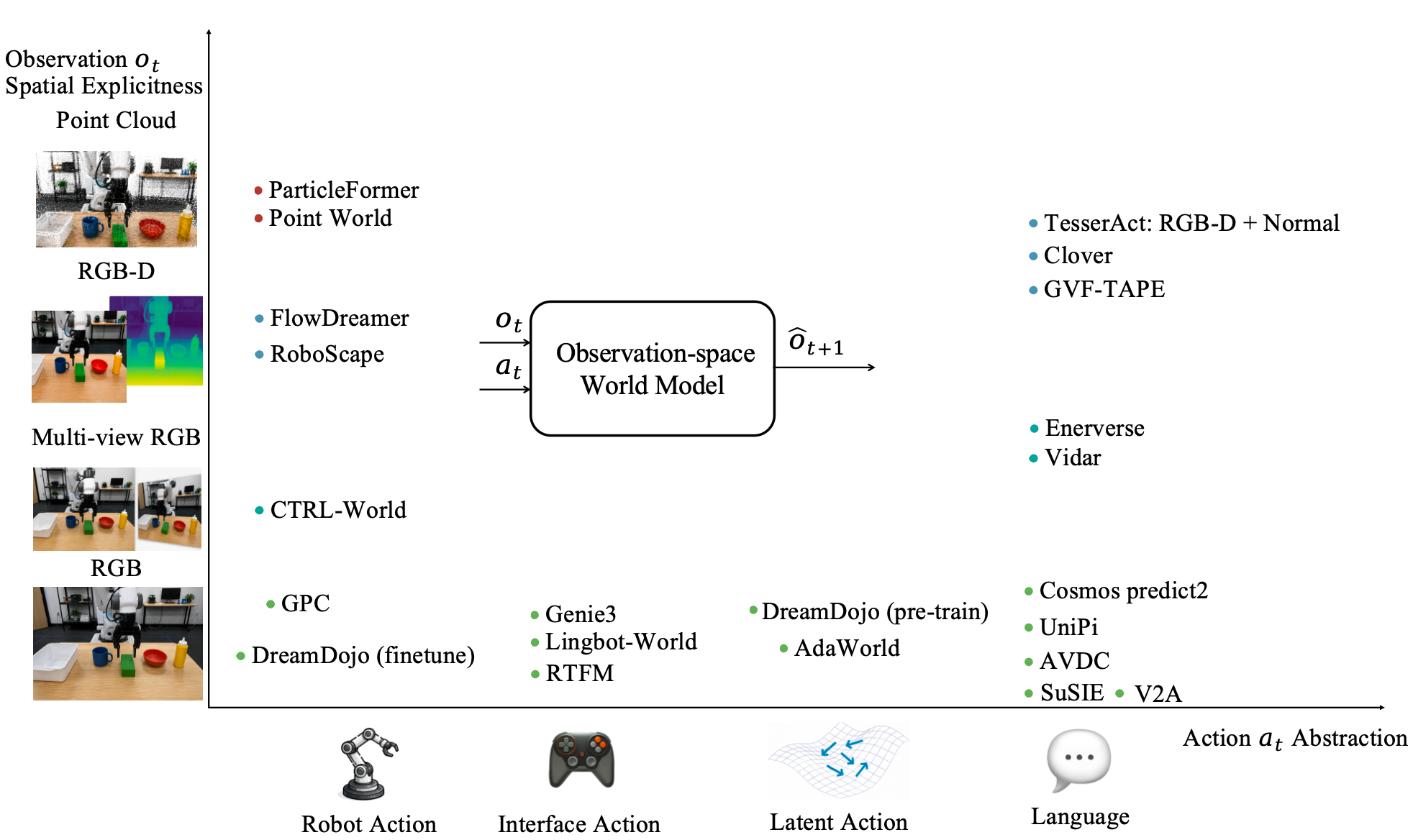}
    \caption{
    Design space of observation-space world models. 
    The vertical axis denotes the spatial explicitness of the observation, ranging from RGB images to multi-view RGB, RGB-D, and point clouds. 
    The horizontal axis denotes the abstraction level of the action conditioning, ranging from low-level robot actions to interface actions, latent actions, and language instructions. 
    Different choices along these two axes lead to different trade-offs between spatial understanding, data availability, and downstream usability.
    }
    \label{fig:obs_space_wm}
\end{figure*}

\section{A Design-Space View of World Models}
\label{sec:design}

While Sec.~\ref{subsec_WM_WAM} shows a world model abstractly as an action-conditioned predictive model, existing world models can be broadly categorized according to the space in which prediction is performed. 
A one-step world model can be written as
$$
y_{t+1} \sim p_\theta(\cdot \mid o_{t}, a_t),
$$
where $p_\theta$ denotes a parameterized predictive model, \(y_{t+1}\) denotes the prediction target, which could be the observation prediction $\hat{o}_{t+1}$ or the state prediction $x_{t+1}$. When $y_{t+1}=\hat{o}_{t+1}$, we refer to the model as an \emph{observation-space} world model. Similarly, when $y_{t+1}=x_{t+1}$, we call it a \emph{state-space world model}. More generally, the prediction target can be a prediction trajectory, such as $y_{t+1:t+H}$.  

% An observation-space world model directly predicts future observations from observation history and actions. 
% In contrast, a state-space world model first encodes observations into an internal state representation, and then predicts the future evolution in that state space. 
% This distinction provides a useful starting point for organizing different world-modeling approaches, since it determines both what the model predicts and how the dynamics are represented.

\subsection{Observation-space World Models}
An observation-space world model predicts future observations directly in the observation space under a given action. 
Given the current observation $o_{t}$ and an action $a_t$, the model estimates the predictive distribution of the next observation:
\[
    \hat{o}_{t+1} \sim p_\theta(\cdot \mid o_{t}, a_t),
\]
where $p_\theta$ is typically parameterized by a neural network. 
Under this formulation, the design space of observation-space world models can be largely characterized by two axes: the type of observation being predicted and the level of abstraction of the action conditioning, as illustrated in Fig.~\ref{fig:obs_space_wm}.

%  {Make classification more clear}
% \paragraph{Classification by observation spatial explicitness.}
% On the observation side, RGB images are the most common representation because they are easy to collect, require inexpensive and widely available sensors, and are abundant in Internet-scale video data ~\cite{gpc,dreamdojo,GoogleDeepMind2026Genie3,lingbot-world,rtfm,adaworld,cosmos,unipi,avdc,susie,v2a}. 
% However, downstream decision-making tasks, especially robotic manipulation, often require the model to capture the spatial structure of the physical scene. 
% This motivates the use of observations with increasing spatial explicitness, including multi-view RGB images ~\cite{ctrl-world,enerverse,feng2025vidar}, RGB-D images ~\cite{flowdreamer,roboscape,tesseract,clover, gvftape}, and point clouds ~\cite{particleformer,pointworld}. 
% These representations provide progressively richer geometric information, but they also introduce a trade-off: spatially explicit data are much less available at scale than ordinary RGB videos. 
% A common compromise is therefore to exploit large-scale RGB video data while enhancing their spatial information through monocular vision models, for example by annotating videos with estimated depth to form pseudo-RGB-D data ~\cite{flowdreamer, roboscape, tesseract, gvftape}.

\paragraph{Observation representation.}
Observation representations can be classified according to how explicitly they encode the spatial structure of a scene: {

\vspace{-3pt}
\noindent \emph{RGB images.}
    RGB images are the most common representation because they are easy to collect, require inexpensive and widely available sensors, and are abundant in Internet-scale video data~\cite{gpc,dreamdojo,GoogleDeepMind2026Genie3,lingbot-world,rtfm,adaworld,cosmos,unipi,avdc,susie,v2a}.
    However, they encode the spatial structure of a scene only implicitly.
    
 \vspace{-3pt}
\noindent \emph{Multi-view RGB images.}
    Multi-view RGB images capture a scene from multiple viewpoints and provide additional geometric constraints for spatial understanding~\cite{ctrl-world,enerverse,feng2025vidar}.
    
 \vspace{-3pt}
\noindent \emph{RGB-D images.}
    RGB-D images augment visual appearance with explicit depth information, making them suitable for spatial reasoning and robotic manipulation~\cite{flowdreamer,roboscape,tesseract,clover,gvftape}. Spatially explicit data are much less available at scale than ordinary RGB videos.
    A common compromise is to annotate large-scale RGB videos with depth estimated by monocular vision models, thereby constructing pseudo-RGB-D training data~\cite{flowdreamer,roboscape,tesseract,gvftape}.
    
    \vspace{-3pt}
\noindent \emph{Point clouds.}
    Point clouds directly represent the three-dimensional geometry of a scene and provide a highly spatially explicit observation representation~\cite{particleformer,pointworld}.
 }

\paragraph{Action representation.}
On the action side, the choice of action representation is closely tied to the intended role of the world model. 
While the observation type determines what form of future is predicted, the action type largely determines how the model can be used. 
A more concrete action representation usually makes the model suitable for control and simulation, whereas a more abstract action representation often shifts the model toward pretraining or planning.

\vspace{-3pt}
\noindent \emph{Robot action.} At the most concrete level, actions correspond to low-level robot commands, such as joint commands, end-effector actions, or continuous control signals ~\cite{particleformer, pointworld, flowdreamer, roboscape, ctrl-world, gpc, dreamdojo}. 
Since these actions are directly grounded in the embodiment of the robot, the resulting world model can predict the physical consequence of executing a specific control command. 
Such models are therefore commonly used as learned simulators for model-predictive control ~\cite{pointworld, particleformer,flowdreamer,gpc}, policy evaluation ~\cite{ctrl-world, roboscape}, or synthetic data generation ~\cite{ctrl-world, roboscape}.

 \vspace{-3pt}
\noindent \emph{Interface action.}  At the next level, interface actions provide a controllable way to interact with the visual world without specifying the low-level motor commands of a particular embodiment. 
For example, given an initial image or scene context, the model may take user-level control signals, camera-control commands, or viewpoint-control inputs, and then synthesize the corresponding future observations ~\cite{GoogleDeepMind2026Genie3, lingbot-world, rtfm}. 
The purpose of this type of world model is not to predict robot dynamics, but to turn a static or partially observed scene into an interactive visual environment. 
It therefore serves as an interface-driven visual simulator, where users or robots can control how the scene is observed and obtain temporally consistent observations under different interactions.

 \vspace{-3pt}
\noindent \emph{Latent action.}  Another line of work introduces latent actions, which are learned from unlabeled videos through reconstruction objectives with an information bottleneck~\cite{dreamdojo,adaworld}. 
Here, the action is not provided by the dataset, but inferred as a compact variable that explains the visual transition between frames. 
This design is mainly motivated by pretraining: by replacing missing action labels with learned latent actions, the world model can exploit large-scale action-free video data and acquire general dynamics priors before being adapted to downstream tasks.

\vspace{-3pt}
\noindent \emph{Language.} At the most abstract level, actions can be represented as language instructions. 
Language-conditioned world models are less grounded in precise low-level control, but they are useful for high-level visual planning. 
Given a task description, such models generate possible future observations that describe how the task may unfold or be completed, thereby providing visual guidance for downstream robots ~\cite{tesseract, clover, enerverse, feng2025vidar, gvftape, cosmos, unipi, avdc, susie, v2a}.

These action representations form a hierarchy ranging from low-level motor control to high-level semantic instructions.

\begin{figure*}[!htbp]
    \centering
    \includegraphics[width=0.9\linewidth]{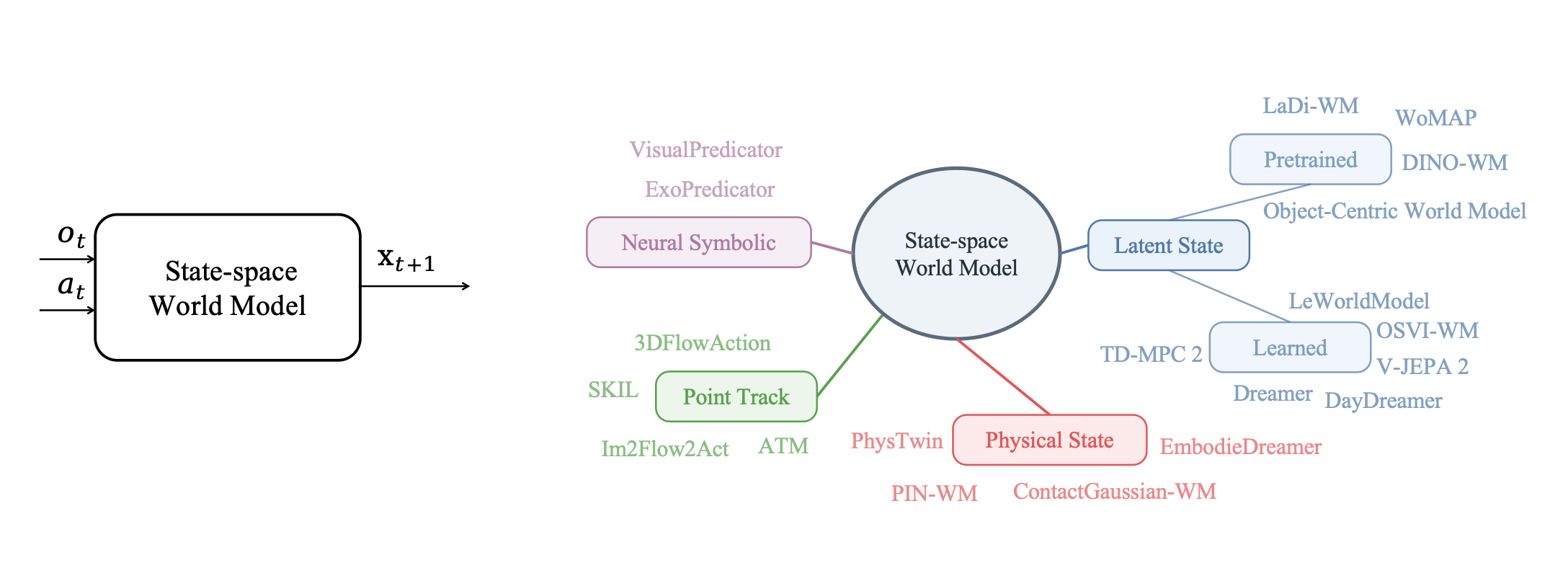 }
    \caption{
    Design space of state-space world models.
    Instead of predicting future observations directly in the raw observation space, state-space world models abstract observations into structured state representations and model their future evolution under actions.
    Representative state choices include latent states, point tracks, neural-symbolic predicates, and physical states.
    Different state representations provide different trade-offs between compactness, semantic structure, physical interpretability, scalability, and downstream usability.
    }
    \label{fig:state_space_wm}
\end{figure*}

\subsection{State-space World Models}

Predicting future observations directly in the observation space is a natural formulation, but it is often difficult to learn because raw observations are high-dimensional and contain many task-irrelevant variations, such as background changes, illumination, texture, or sensor noise. As a result, modeling dynamics directly in the observation space may force the model to explain unnecessary visual details rather than the task-relevant evolution of the world. State-space world models address this issue by first abstracting observations into a more compact and structured state representation, and then predicting the future evolution in this state space:
\[
x_{t+1} \sim p_{\theta}(\cdot \mid o_{t}, a_t),
\]
where $x_{t+1}$ denotes the state prediction of interest, which may be a latent vector, a set of tracked points, a collection of symbolic predicates, or physical state variables. Unlike observation-space world models, the transition model $p_\theta$ is not necessarily restricted to a neural network. Depending on the choice of state representation, the transition model may be implemented as a neural network, a symbolic or neural-symbolic operator model, or a physics-based simulator. By predicting structured states instead of raw observations, state-space world models reduce the complexity of dynamics learning and focus the prediction on task-relevant or physically meaningful aspects of the world. As illustrated in Fig.~\ref{fig:state_space_wm}, existing state-space world models can be organized according to the type of state representation they use.

\paragraph{Latent state model.} A major class of state-space world models uses latent states, where observations are encoded into latent vectors by neural networks ~\cite{huang2025ladiwm,yin2025womap,jeong2025ocwm,zhou2024dinowm,maes2025leworldmodel,goswami2025osviwm,assran2025vjepa2,wu2022daydreamer,hafner2023dreamerv3,hansen2024tdmpc2}. These latent states can be obtained in different ways. Some methods learn the latent representation directly from the target domain, for example through reconstruction-based objectives ~\cite{wu2022daydreamer,hafner2023dreamerv3} or through predictive objectives that model future latent states without reconstructing pixels ~\cite{goswami2025osviwm,maes2025leworldmodel,assran2025vjepa2,hansen2024tdmpc2}. Such representations are usually adapted to the specific environment, task distribution, and control setting. Other methods use pretrained visual or vision-language models to extract latent features as state representations ~\cite{huang2025ladiwm,yin2025womap,jeong2025ocwm,zhou2024dinowm}. Since these encoders are trained on large-scale Internet data, the resulting states often provide more general semantic and visual abstractions. Latent-state world models are commonly paired with robot actions and are often used as compact learned simulators for reinforcement learning, or model-predictive control.

\paragraph{Point track model.}  Another line of work represents the state as point tracks ~\cite{zhi20253dflowaction, wang2025skil, xu2024im2flow2act, wen2024atm}. These points can be two-dimensional ~\cite{wang2025skil, xu2024im2flow2act, wen2024atm} or three-dimensional ~\cite{zhi20253dflowaction}, randomly sampled ~\cite{wen2024atm} or semantically selected ~\cite{xu2024im2flow2act, zhi20253dflowaction, wang2025skil}, and their trajectories describe how task-relevant parts of the scene move over time. Compared with raw image prediction, point-track prediction removes much of the irrelevant background variation while preserving explicit motion information in the scene. This makes the future representation more compact and structured, without requiring the model to generate full visual observations. However, the choice of points also introduces an additional structural prior: although this prior can simplify learning, it may reduce scalability when the relevant scene elements are hard to define or vary significantly across tasks. In many applications, point-track world models are conditioned on language instructions and predict how key points in the scene should move in order to accomplish the task. The predicted tracks can then be used as reference trajectories or intermediate goals for a downstream controller.

\vspace{-2pt}

\paragraph{Neural-symbolic model.} Neural-symbolic world models use a set of grounded predicates as the state representation ~\cite{liang2025exopredicator, liang2024visualpredicator}. A neural-symbolic predicate is a symbolic Boolean fact whose truth value is grounded in raw perception by a neural model. This representation maps continuous visual observations into a compact set of logical facts that can support reasoning and planning. In this setting, actions are usually high-level skills rather than low-level motor commands. The transition model describes the preconditions and effects of these skills over the predicate state, similar in spirit to symbolic planning models such as PDDL, but with predicates grounded in perception. The goal of this type of world model is to learn a compact symbolic abstraction for long-horizon planning. By representing skills through their preconditions and effects, neural-symbolic world models support compositional reasoning, efficient search, and interpretable planning over high-level task structures.

\paragraph{Physical state model.} A further class of state-space world models is based on physical states and classical mechanics ~\cite{li2025pinwm,jiang2025phystwin,wang2025embodiedreamer,wang2025contactgaussianwm}. Instead of learning arbitrary latent dynamics from data, these methods explicitly represent the world using physical variables such as object poses, velocities, contact states, masses, friction coefficients, or other dynamics parameters. This formulation incorporates strong human priors about how the physical world evolves, and is closely related to the modeling assumptions used in physics simulators. A typical pipeline first reconstructs the target scene in 3D, then aligns a simulated environment with the real scene by replaying real interaction trajectories and optimizing dynamics and rendering parameters. Once the simulator is aligned, future states can be predicted by solving physics-based transition equations conditioned on the current physical state and the robot action. Such world models are especially useful for model-predictive control, reinforcement learning, and synthetic data generation, where accurate physical rollouts are more important than directly generating raw visual observations.

\section{World Action Models}
\label{sec_WAM}

\begin{figure*}[!htbp]
\centering
\includegraphics[width=0.95\linewidth]{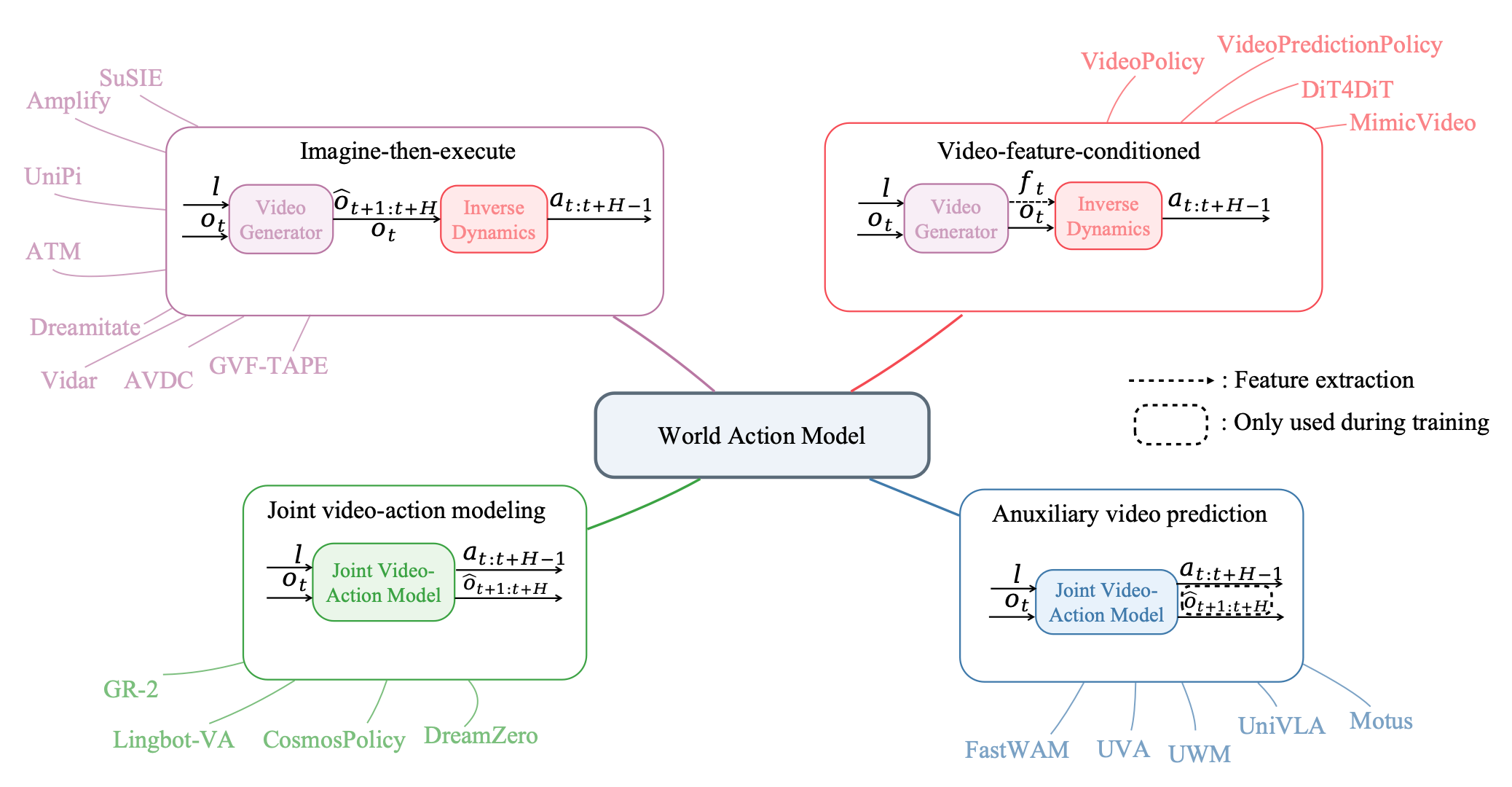}
\caption{
Taxonomy of world action models.
Given the observation $o_t$ and language instruction $l$, world action models couple future observation prediction with robot action generation in different ways.
Representative paradigms include imagine-then-execute, video-feature-conditioned action prediction, joint video-action modeling, and auxiliary video prediction for policy learning.
}
\label{fig:wam}
\end{figure*}

The previous section categorizes world models according to their prediction space. 
We focus on video-based world action models that incorporate future-oriented visual modeling in one of four ways: explicit future generation, conditioning on predictive video features, joint video-action generation, or auxiliary future prediction during policy training. In physical decision making, prediction alone 
is insufficient; the robot must also infer executable actions that can realize the predicted 
future. This motivates world action models, which couple visual future prediction with 
action generation:
\[
(\hat{o}_{t+1:t+H}, a_{t:t+H-1}) \sim p_\psi(\cdot \mid o_t, l).
\]
Existing approaches differ in how this coupling is implemented, leading to four 
representative paradigms summarized in Fig.~\ref{fig:wam}.

\paragraph{Imagine-then-execute.}
The most direct paradigm is a two-stage \emph{imagine-then-execute} formulation ~\cite{unipi,susie,wen2024atm,liang2024dreamitate,avdc,feng2025vidar,gvftape,RIGVid}. 
In the first stage, a video prediction model generates visual subgoals:
\[
\hat{o}_{t+1:t+H} \sim p_\theta(\cdot \mid o_t, l).
\]
In the second stage, a separate inverse dynamics model or goal-conditioned policy, denoted by the parameterized distribution $q_\phi$, maps the current observation and a predicted visual subgoal into robot actions,
\[
a_{t:t+H-1} \sim q_\phi(\cdot \mid o_t, \hat{o}_{t+1:t+H}).
\]
The two modules together form a world action model: the video model imagines a possible future, and the inverse dynamics model grounds this future into executable actions. This decomposition has several practical advantages. Since the visual planner and the action grounding model are separated, they can be trained on different data sources: the video model can benefit from large-scale visual data, while the inverse dynamics model can be trained on robot trajectories with action labels. The modular design also allows additional structure to be injected into the action grounding process. For example, one can estimate intermediate quantities such as end-effector poses ~\cite{gvftape, liang2024dreamitate}, object poses ~\cite{RIGVid}, or dense optical flow ~\cite{avdc} from the generated video, and then use these cues to derive or guide the corresponding robot actions. This decomposition provides flexibility, but it also makes action prediction dependent on the quality of the generated visual subgoals. Errors in the predicted subgoals may propagate to the inverse dynamics model, causing the predicted actions to deviate from the intended task execution.

\paragraph{Video-feature-conditioned action prediction.}
A second paradigm keeps the two-stage structure but does not require the video model to generate a complete future observation sequence ~\cite{videopolicy,hu2025vpp,ma2026dit4dit,pai2026mimicvideo,wang2026ssi}. This design is motivated by the computational cost of diffusion-based models, where producing full future frames may require multiple iterative sampling steps. Instead of decoding a complete video rollout, these methods extract intermediate spatiotemporal features from the video prediction model and use them to condition an action prediction module:
\[
f_t = \mathcal{H}\big(u_\theta(o_t, l)\big),
\qquad
 \sim q_\phi(\cdot \mid o_t, f_t),
\]
where $u_\theta$ denotes the neural backbone of the video prediction model, $\mathcal{H}(\cdot)$ denotes a feature extraction operation, and $f_t$ denotes the extracted spatiotemporal representation. The underlying assumption is that the internal representations of a video model already contain useful information about task intent, scene dynamics, and possible future evolution, even if the model does not explicitly decode the final video during policy inference. This can substantially reduce inference cost compared with full video generation, while still transferring predictive visual representations to the action model. The main limitation is that the interface between the video model and the policy becomes a latent feature rather than an explicit visual plan. As a result, the intermediate representation is less interpretable, harder to inspect, and may not expose whether the policy is using meaningful future information or merely exploiting task-correlated features.

\paragraph{Joint video-action modeling.}
A third paradigm attempts to model future observations and robot actions jointly within a single generative model ~\cite{ye2026dreamzero,kim2026cosmospolicy,cheang2024gr2,li2026lingbotva}:
\[
(\hat{o}_{t+1:t+H}, a_{t:t+H-1})
\sim p_\theta(\cdot \mid o_t, l).
\]
In this formulation, video prediction and action prediction are not treated as two separate modules. Instead, the model learns a joint distribution over visual futures and the corresponding action sequences. A common strategy is to start from a video generation backbone pretrained on large-scale visual data ~\cite{cosmos,wan2025}, modify the architecture or output space so that it can also produce robot actions, and then adapt the model on robot trajectories with action labels. This unified formulation has the advantage that future observations and actions are learned in a shared representational space, which can improve their consistency: the generated actions are trained together with the visual future they are supposed to induce. However, joint modeling also inherits the difficulties of both video generation and action learning. It requires robot data with action labels for adapting the action output, and such data are much scarcer than ordinary videos. In addition, the model must jointly handle high-dimensional visual generation and precise action prediction, whose optimization objectives may not always be aligned. 

\paragraph{Auxiliary video prediction for policy learning.}
A fourth paradigm uses video prediction as an auxiliary training objective rather than as an explicit inference-time module~\cite{yuan2026fastwam,zhu2025uwm,li2025uva,bu2025univla,bi2026motus}. In this setting, the policy is trained with an additional prediction branch that reconstructs or generates future observations from the policy's internal representation. The purpose of this branch is not to produce an explicit visual plan for execution, but to encourage the policy backbone to encode spatiotemporal information related to future scene evolution. During training, the video prediction loss provides an auxiliary supervisory signal in addition to the action prediction loss. During inference, the auxiliary video branch can be removed, and only the action prediction head is used. This design avoids the cost of generating videos at test time while still using future prediction to shape the learned representation. Its limitation is that the predicted future is not explicitly used as a plan during execution. Therefore, the benefit of the video objective depends on whether the auxiliary prediction task induces representations that are useful for action selection. Balancing the action loss and the video prediction loss can also be nontrivial, since accurate visual prediction does not necessarily lead to better control.

Overall, these four paradigms represent different ways of coupling world prediction with action generation. Imagine-then-execute methods provide modularity and interpretability, but may suffer from a visual-action grounding gap. Feature-conditioned methods reduce the computational cost of explicit video generation, but sacrifice the transparency of visual plans. Joint video-action models offer a unified formulation, but require action-labeled robot data and must jointly optimize visual and action prediction. Auxiliary prediction methods improve efficiency at inference time, but rely on the training objective to implicitly transfer predictive information into the policy. The central design trade-off is therefore how explicitly the future should be represented and how tightly this future representation should be coupled with action prediction. In addition, \cite{zheng2026video} provide an alternative perspective that  categorizes video-based manipulation methods according to the interface through which temporal visual information is converted into control.


\bibliography{reference}
\bibliographystyle{icml2025}

% %%%%%%%%%%%%%%%%%%%%%%%%%%%%%%%%%%%%%%%%%%%%%%%%%%%%%%%%%%%%%%%%%%%%%%%%%%%%%%%
% %%%%%%%%%%%%%%%%%%%%%%%%%%%%%%%%%%%%%%%%%%%%%%%%%%%%%%%%%%%%%%%%%%%%%%%%%%%%%%%
% % APPENDIX
% %%%%%%%%%%%%%%%%%%%%%%%%%%%%%%%%%%%%%%%%%%%%%%%%%%%%%%%%%%%%%%%%%%%%%%%%%%%%%%%
% %%%%%%%%%%%%%%%%%%%%%%%%%%%%%%%%%%%%%%%%%%%%%%%%%%%%%%%%%%%%%%%%%%%%%%%%%%%%%%%
% \newpage
% \newpage

% \appendix
% \onecolumn

% You can have as much text here as you want. The main body must be at most $8$ pages long.
% For the final version, one more page can be added.
% If you want, you can use an appendix like this one.  

% The $\mathtt{\backslash onecolumn}$ command above can be kept in place if you prefer a one-column appendix, or can be removed if you prefer a two-column appendix.  Apart from this possible change, the style (font size, spacing, margins, page numbering, etc.) should be kept the same as the main body.
% %%%%%%%%%%%%%%%%%%%%%%%%%%%%%%%%%%%%%%%%%%%%%%%%%%%%%%%%%%%%%%%%%%%%%%%%%%%%%%%
% %%%%%%%%%%%%%%%%%%%%%%%%%%%%%%%%%%%%%%%%%%%%%%%%%%%%%%%%%%%%%%%%%%%%%%%%%%%%%%%

\end{document}